\documentclass[sigconf,nonacm]{acmart}
\usepackage[utf8]{inputenc}
\usepackage[T1]{fontenc}
\usepackage{algorithmic}
\usepackage[ruled,vlined]{algorithm2e}
\usepackage{soul} 

\usepackage{caption}
\usepackage{amsmath}
\DeclareMathOperator*{\argmax}{arg\,max}

\usepackage[ruled,vlined]{algorithm2e}
\usepackage{soul} 
\usepackage{array}
\usepackage{tabularx}

\usepackage{tkz-graph}
\usetikzlibrary{shapes.geometric}

\usepackage{tikz}
\usetikzlibrary{decorations.pathreplacing}

\newtheorem{definition}{definition}

\usepackage{mathtools}
\DeclarePairedDelimiter\floor{\lfloor}{\rfloor}

\usepackage[position=bottom]{subfig}
\AtBeginDocument{%
  \providecommand\BibTeX{{%
    \normalfont B\kern-0.5em{\scshape i\kern-0.25em b}\kern-0.8em\TeX}}}

\copyrightyear{2022}
\acmYear{2022}
\setcopyright{acmcopyright}\acmConference[GECCO '22 Companion]{Genetic and Evolutionary Computation Conference Companion}{July 9--13, 2022}{Boston, MA, USA}
\acmBooktitle{Genetic and Evolutionary Computation Conference Companion (GECCO '22 Companion), July 9--13, 2022, Boston, MA, USA}
\acmPrice{15.00}
\acmDOI{10.1145/3520304.3533994}
\acmISBN{978-1-4503-9268-6/22/07}

\begin{document}
\begin{small}
  \title{\vspace*{-1.0cm}Towards QD-suite: developing a set of benchmarks for Quality-Diversity algorithms}

\author{Achkan Salehi and Stephane Doncieux}
\affiliation{%
  \institution{Sorbonne Université, CNRS, ISIR, F-75005}
  \city{Paris} 
  \country{France} 
}
\email{{achkan.salehi,stephane.doncieux}@sorbonne-universite.fr}

 \begin{abstract}
   While the field of Quality-Diversity (QD) has grown into a distinct branch of stochastic optimization, a few problems, in particular locomotion and navigation tasks, have become \textit{de facto} standards. Are such benchmarks sufficient? Are they representative of the key challenges faced by QD algorithms? Do they provide the ability to focus on one particular challenge by properly disentangling it from others? Do they have much predictive power in terms of scalability and generalization? Existing benchmarks are not standardized, and there is currently no MNIST equivalent for QD. Inspired by recent works on Reinforcement Learning benchmarks, we argue that the identification of challenges faced by QD methods and the development of targeted, challenging, scalable but affordable benchmarks is an important step. As an initial effort, we identify three problems that are challenging in sparse reward settings, and propose associated benchmarks: (1) \textit{Behavior metric bias}, which can result from the use of metrics that do not match the structure of the behavior space. (2) \textit{Behavioral Plateaus}, with varying characteristics, such that escaping them would require adaptive QD algorithms and (3) \textit{Evolvability Traps}, where small variations in genotype result in large behavioral changes. The environments that we propose satisfy the properties listed above.
 \end{abstract}




\maketitle

\section{Introduction}

Quality-Diversity (QD) \cite{cully2017quality} algorithms have become the subject of increasing interest in the research community, mainly due to their reported success in many areas ranging from Reinforcement Learning (RL) and Control \cite{pierrot2020diversity, kim2021exploration, cully2015robots} to vehicle conception \cite{gaier2018data} and game level design \cite{charity2020baba}. As QD has arguably grown into a distinct branch of stochastic optimization \cite{chatzilygeroudis2021quality}, a few problems, such as locomotion \cite{cully2015robots, mouret2020quality} or navigation tasks \cite{sfikas2021monte, vassiliades2017comparison, gravina2016surprise} (and sometimes combinations of the two \cite{paolo2021sparse, colas2020scaling}) have repeatedly been used, becoming \textit{de facto} benchmarks. To what extent are these evaluation environments sufficient, or even relevant? Are they really representative of the general challenges that QD methods might face? Do they have much predictive power in terms of generalization to other environments? At the very least, they are not standardized, and none of them can be seen as \texttt{MNIST} or \texttt{CIFAR-10} equivalents for QD. Common benchmarks are important tools for structuring scientific communities and helps them make faster progress \citep{sim2003using}. It is a sign of maturity that allows to enter into Kuhn's \textit{normal science} \citep{kuhn1970ttie}. 

Motivated by the above, we draw inspiration from \texttt{bsuite} \cite{osband2019behaviour} which proposes a set of benchmarks for RL. Of particular interest to us are the desiderata identified by the authors, according to which a benchmark should be:

\begin{enumerate}
  \item \textbf{Targeted:} it should correspond to a key problem faced by the community.
  \item \textbf{Simple:} it should strip away confounding/confusing factors in research.
  \item \textbf{Challenging:} the problem must encourage the development of methods that expand the current boundaries.
  \item \textbf{Fast:} Execution should be possible on standard hardware and allow quick iterations (ideally, multiple iterations through a day, from launch to results).
  \item \textbf{Scalable:} the benchmark should provide insight on scalability, not performance on one environment.
\end{enumerate}

It should be noted that QD algorithms \textemdash aside from having applications outside of RL \textemdash are fundamentally different from most forms of policy search as their main goal is to generate sets of solutions that cover a behavior space. This frees QD algorithms from some of the typical optimization problems such as premature convergence to local minima. Furthermore, as we will discuss throughout the paper, QD algorithms face challenges of their own. Therefore, while existing RL environments are indeed valuable tools for evaluating QD algorithms, they should not be the only ones. Our aim in this paper is to encourage the development of a \texttt{QD-suite}: a benchmark tailored to the specificities of QD, that would embody the five properties enumerated above.

As an initial effort in this direction, we report three challenges that we have identified in sparse/non-existent reward settings, and for each case, propose simple environments that isolate those problems:

\begin{itemize}
  \item \textbf{The behavior metric bias}. Situations in which the distance function does not capture the structure of the reachable behavior space can lead to selection pressures that prevent the population from reaching certain parts of the search space. The benchmark that we present to highlight that problem (\S\ref{sec_spiral}) shows that without the use of the proper geodesic distance and compensating for bias inducing mutation schemes, QD algorithms, even with unreasonably large archives, can struggle to cover the entire search space.
  \item \textbf{Behavioral plateaus.} The ability of QD algorithms in "escaping" areas of the genotype space over which the behavior function is constant (or where evolvability is very low) is dependent on their underlying mutation and selection schemes. The benchmark problem that we present is composed of various such plateaus, and suggests that QD algorithms should adapt themselves to the structure of the behavior space.
  \item \textbf{Evolvability Traps.} The third challenge that we have identified is related to areas of the behavior space in which small variations in genotype create large changes in behavior. QD methods can be lured by these areas at the expense of subspaces that would have been more promising in the longer term.
\end{itemize}

This list of is clearly not exhaustive, and is expected to grow as new challenges are identified in the future.

The environments that we propose are simple, and each of them focuses on a different problem with which QD methods struggle. They are suitable for rapid experimentation without requiring powerful clusters, as they do not require costly simulations. Additionally, the difficulties that they highlight can occur in more complex problems at larger scales. They are thus well aligned with the qualities that we listed.

In the next section, the notations that will be used throughout the paper, as well as the reference methods on which the benchmarks have been tested will be introduced. The following three sections (\S\ref{sec_spiral}, \S\ref{section_ssf} and \S\ref{section_deceptive_evo}) introduce benchmarks associated to the three proposed challenges. 

\section{Notations and Reference methods} 

Throughout the paper, genotype spaces will be noted $\mathcal{G}$. The behavior descriptor space and the behavior function will respectively be noted $\mathfrak{B}$ and $\phi:\mathcal{G} \rightarrow \mathfrak{B}$. When necessary, various superscripts and subscripts will be introduced in order to differentiate between different parametrizations of genotypes, behaviors and behavior mappings. The population at generation $g$ will be noted $\mathcal{P}^g$, and we will write $\mathcal{G}_{max}$ the maximum number of generations.

\textbf{Reference methods.} For the first benchmark, experiments were based on the Novelty Search algorithm \cite{lehman2011abandoning} with both \textit{unstructured} and \textit{structured} archives. In the latter case, we also experimented with resampling in the archive, making the method a hybrid between NS and MAP-elites \cite{mouret2015illuminating}. For the second and third benchmarks, we report results from evaluating both NS with an unstructured archive and MAP-elites using a vanilla grid-based archive. In the case of unstructured archives with bounded sizes, individuals added/removed to/from the archive are selected randomly.

\section{Behavior distance bias: Archimedean-Spiral-v0}
\label{sec_spiral}

The aim of the toy environments that we develop in this section is to show that the non-linearity of the behavior function $\phi$ as well as the use of inadequate distance metrics \footnote{In general, QD algorithms \textemdash whether they belong the NS or MAP-elites family \textemdash use the Euclidean distance by default. Some exceptions exist, such as in robotic manipulation where distances are measured in the tangent plane of $SE(N)$ \cite{salehi2021few}. However, those also do not correspond to the true structure of the reachable behavior space, which is defined by the scene and the limits of the actuators.} results in biased exploration, which can be difficult to overcome by common QD algorithms.

Let us assume a reachable behavior space $\mathfrak{B}_{spiral}$ given by an archimedean spiral (figure \ref{fig_spiral_setup}) embedded in $\mathbb{R}^2$ whose equation in Cartesian coordinates is given by (for $t \in \mathbb{R}^{+}$)

\begin{equation}
  \gamma(t)=(at\cos(t), at\sin(t))
\end{equation}

and for which the geodesic distance naturally coincides with its arc-length, given by

\begin{equation}
  S(t_1,t_2)=\frac{a}{2}(t\sqrt{t^2+1} + \log(t+\sqrt{t^2+1}))\rvert_{t_1}^{t_2}.
\end{equation}

In general, the geodesic distance is unknown to the QD algorithm. Instead, the Euclidean metric from the $\mathbb{R}^N$ space in which the behavior space is embedded is used.

Let us now define the two distinct bounded genotype spaces 

\begin{equation}
  \begin{cases}
    \mathcal{G}_b\triangleq [0,\alpha\pi]\\
    \mathcal{G}_u\triangleq \{S(0,l) \rvert l \in [0, \alpha\pi]\}
  \end{cases}
\end{equation}

for some positive real $\alpha$. Indeed, $\mathcal{G}_b$ corresponds to the angle parametrization of the curve, and $\mathcal{G}_u$ is the set of possible arc-lengths. Let us use an isotropic Gaussian mutation operator, and finally, let us define two mappings from those genotypes to $\mathbb{R}^2$:

\begin{equation}
  \begin{cases}
    \phi_b(g_b)\triangleq \gamma(g_b) & \forall g_b \in \mathcal{G}_b \\
    \phi_u(g_u)\triangleq \gamma(S^{-1}(g_u)) & \forall g_u \in \mathcal{G}_u
  \end{cases}
\end{equation}. 

The motivation behind this choices for $\phi_b$ and $\phi_u$, which both have $\mathfrak{B}_{spiral}$ as their image, is that they behave differently under Gaussian mutations: it can easily be verified that an isotropic Gaussian distribution in $\mathcal{G}_b$ will be mapped by $\phi_b$ to a distribution that is skewed towards the exterior of the spiral, while an isotropic Gaussian distribution in $\mathcal{G}_u$ will be mapped by $\phi_u$ to an isotropic Gaussian on $\gamma$. Note that the mapping $S^{-1}$ can not be expressed in closed form. As a result, whenever necessary, we approximate its values by solving the corresponding ordinary differential equation.

All experiments in that section were based on NS and NS/MAP-elites hybrids resulting from pairing NS with a structured archive and/or allowing resampling in the archive instead of in the active population. We set $\alpha=30$, with population and offspring sizes of $M=N=30$, mutations sampled from $\mathcal{N}(0,0.3)$ and considered $k=10$ nearest neighbors to compute the NS objective. In all of the experiments, the number of generations was set to $G_{max}=1000$, and all individuals were initialized with a value of $s$ (the red dot in figure \ref{fig_spiral_setup}).

Consider the four possible problems defined by pairing a distance function with a parametrization. It can be empirically verified (figure \ref{spiral_metrics}) that without the use of an archive, NS is unlikely to cover the search space unless both the correct geodesic distance is used in conjunction with the unbiased parametrization $\phi_u$. 

As knowing/learning the correct geodesic distance and the availability of mutation operators that do not induce local bias are impractical in more complex problems, we focus from now on on the case were the Euclidean metric is used with the bias-inducing parametrization $\mathcal{G}_b$.

Interestingly, in that situation, NS still struggles when endowed with archives of considerable size (relative to problem dimensionality and the size of the behavior space). Figure \ref{spiral_metrics_archive} shows the cumulative results (across runs) of NS experiments with both unstructured and structured archives, when the Euclidean metric is used with the bias-inducing parametrization $\mathcal{G}_b$. As figures \ref{spiral_metrics_archive} (a, b) demonstrate, even unreasonably large archive sizes \footnote{Recall that the time complexity of NS is $O(N\log N)$ with $N$ the archive size\cite{mouret2015illuminating}. Thus, in practical applications, it is desirable to bound the archive. The inability of a method to solve a simple low-dimensional problem using a small archive might hint at scalability issues.} do not lead to complete behavior space coverage. We found that in order to ensure full coverage consistently, the archive needed to grow very large (>$1000$ individuals, see figure \ref{spiral_metrics_archive}(c)). Note that sampling from the archive instead of only from the active population (figure \ref{spiral_metrics_archive} (e)) did not result in significant improvements.

Structured archives (in the spirit of vanilla MAP-elites \cite{mouret2015illuminating}) which also require some effort in order to tune their resolution to the behavior space, did not lead to success either \ref{spiral_metrics_archive} (d, f). The main reason for that outcome is that once the exterior of the spiral has been covered, the discovery of the parts towards the center depends on the selection of a single particular cell, which can not be done in a reasonable time without biasing the selection. 

As discussed in \cite{salehi2021few}, a potential solution to that problem is the use of complementary selection pressures: in particular, a promising solution could result from combining NS with selection pressures that are \textit{absolute} (as opposed to relative to an ever-moving archive), such as the occupancy information that is made available from the use of a structured archive.

\begin{figure}[ht!]
  \includegraphics[width=50mm,clip]{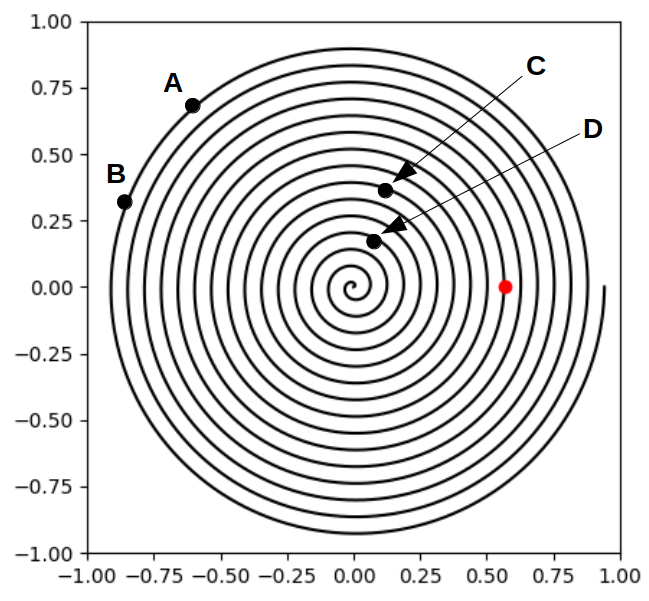}\\
  \caption{An Archimedean spiral with parameter $a=0.01$ and $t \in [0, 30\pi]$ embedded in $\mathbb{R}^2$. It is clear that using the Euclidean distance and $k=1$, points $A$ and $B$ will have higher Novelty scores than $C$ and $D$. In contrast, using the geodesic distance, $C$ and $D$ would have much higher Novelty than $A$ and $B$. In this figure, the red dot indicates the starting point of the agents in our experiments.}
  \label{fig_spiral_setup}
\end{figure}

\begin{figure}[ht!]
  \centering
  \captionsetup[subfigure]{justification=centering}
    \subfloat[Euclidean distance, $\phi_b$ parametrization.][Euclidean distance,\\ $\phi_b$ (angle) parametrization.]{
      \includegraphics[width=42mm,trim={1cm 0 1cm 1.4cm},clip]{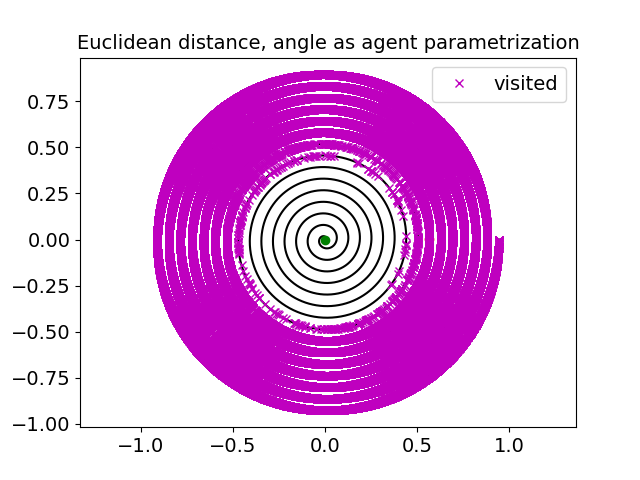}}
    \subfloat[Euclidean distance, $\phi_u$ parametrization.][Euclidean distance,\\ $\phi_u$ (arc-length) parametrization.]{
      \includegraphics[width=42mm,trim={1cm 0 1cm 1.4cm},clip]{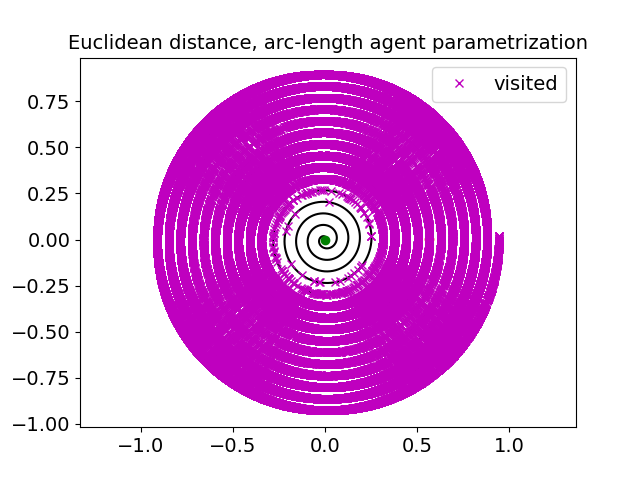}}\\
    \subfloat[Geodesic distance, $\phi_b$ (angle) parametrization.][Geodesic distance,\\ $\phi_b$ (angle) parametrization.]{
      \includegraphics[width=42mm,trim={1cm 0 1cm 1.4cm},clip]{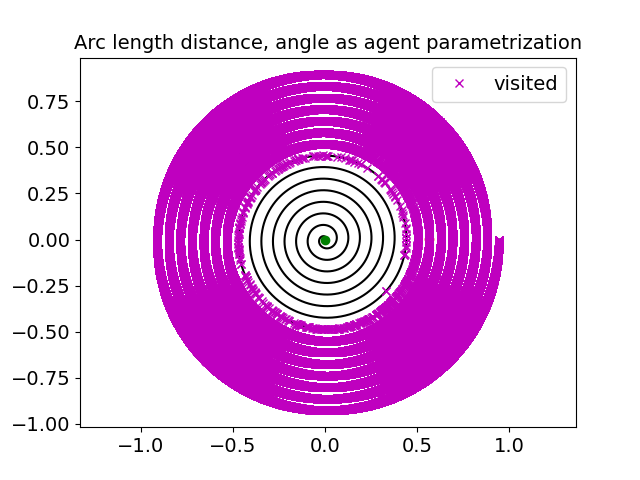}}
    \subfloat[Geodesic distance, $\phi_u$ (arc length) parametrization.][Geodesic distance,\\ $\phi_u$ (arc length) parametrization.]{
      \includegraphics[width=42mm,trim={1cm 0 1cm 1.4cm},clip]{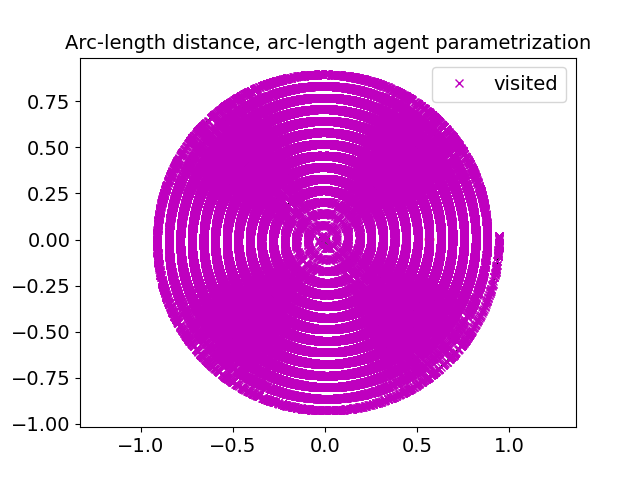}}\\
    \caption{The cumulative results of $20$ Archive-less Novelty search experiments in the four problem settings. Note that in (a) and (c), a Gaussian distribution centered at $g_b \in \mathcal{G}_b$ is mapped to a distribution with mean $\phi_u(g_b)$ but that is skewed towards the exterior of the spiral. In (a) and (b), the use of the Euclidean metric results in lower Novelty scores towards the interior of the spiral.}
   \label{spiral_metrics}
\end{figure}

\begin{figure}[ht!]
  \centering
  \captionsetup[subfigure]{justification=centering}
    \subfloat[Unstructured archive, maximum size of $100$][Unstructured archive,\\ maximum size of $100$]{
    \includegraphics[width=40mm,trim={0.6cm 0 1.0cm 1.4cm},clip]{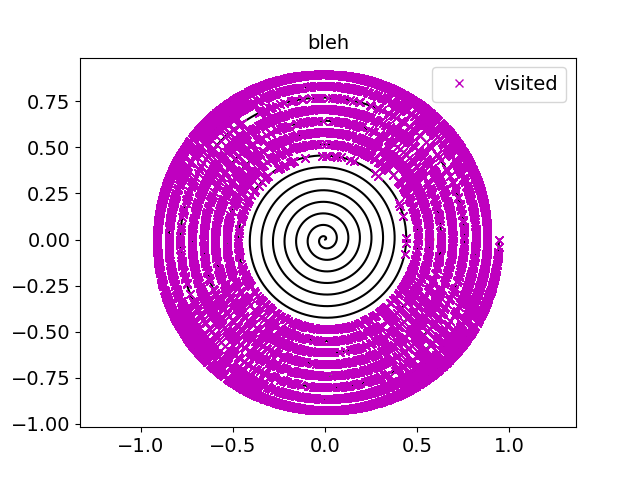}}
    \subfloat[Unstructured archive, maximum size of 200][Unstructured archive,\\ maximum size of 200]{
    \includegraphics[width=40mm,trim={1cm 0 1cm 1.4cm},clip]{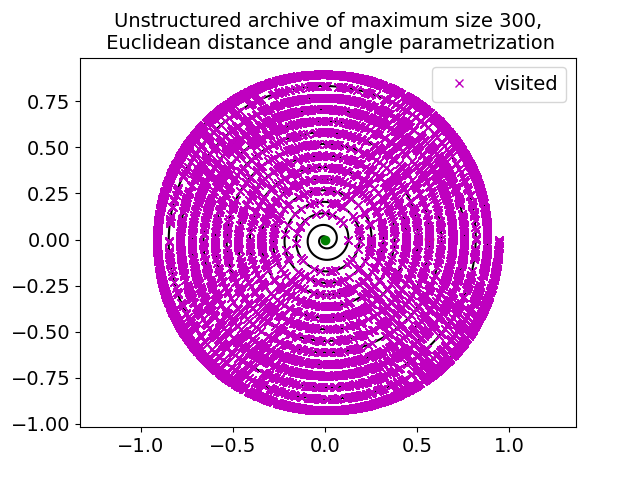}}
  
  \subfloat[Unstructured archive, maximum size of 3000][Unstructured archive,\\ maximum size of 3000]{
    \includegraphics[width=40mm,trim={1cm 0 1cm 1.4cm},clip]{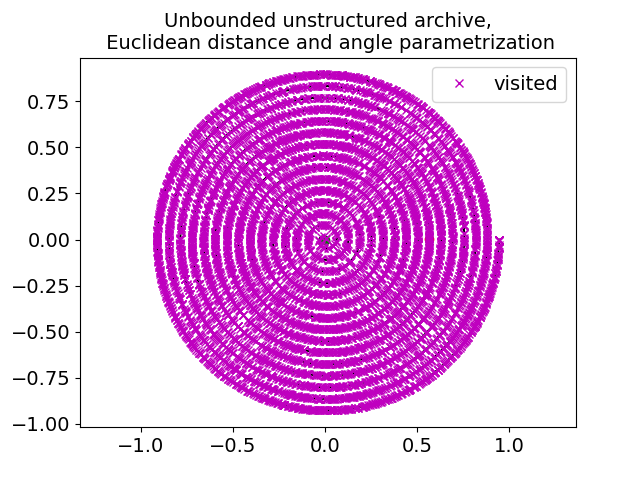}}
  \subfloat[Structured archive,no resampling in the archive][Structured archive,\\ no resampling in the archive]{
    \includegraphics[width=40mm,trim={1cm 0 1cm 0},clip]{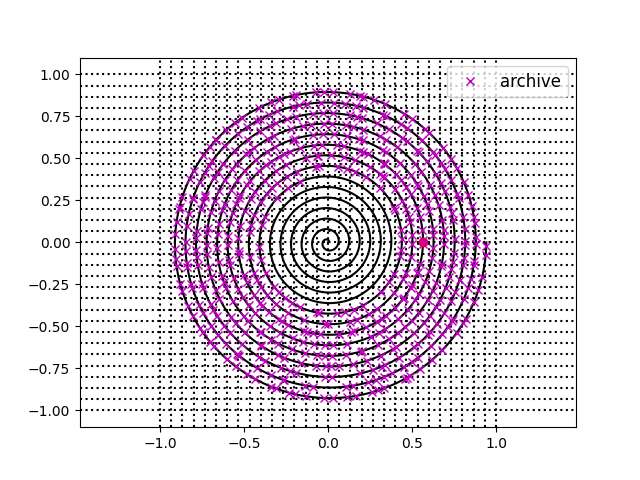}}
  
  \subfloat[Unstructured archive, maximum size of 200 and resampling in the archive][Unstructured archive,\\ maximum size of 200 and\\ random resampling in the archive]{
    \includegraphics[width=40mm,trim={1cm 0 1cm 1.4cm},clip]{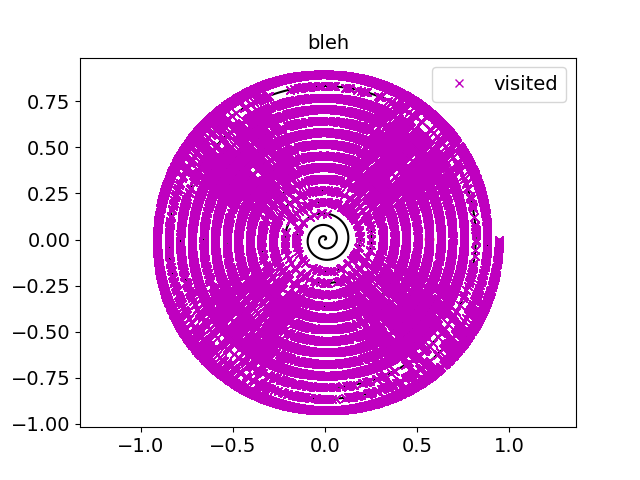}}
     \subfloat[Structured archive,random resampling in the archive][Structured archive,\\ random resampling in the archive]{
    \includegraphics[width=40mm,trim={1cm 0 1cm 0},clip]{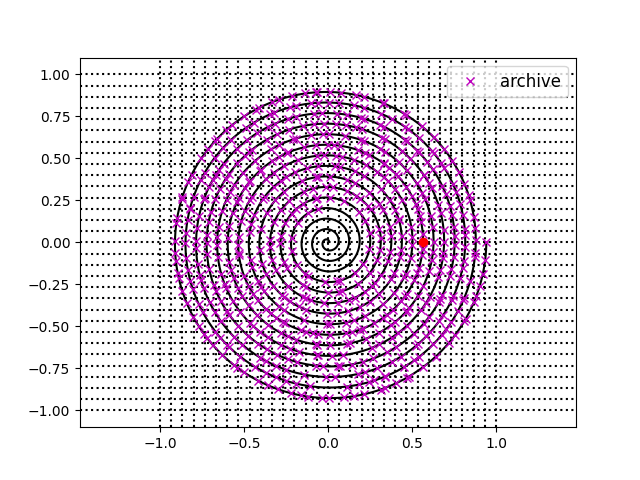}}
  \caption{Cumulated results of experiments (with $G_{max}=1000$) obtained with different archive configurations when the Euclidean metric and the $\phi_b$ parametrization were used.}
   \label{spiral_metrics_archive}
\end{figure}

\section{Behavioral plateaus: SSF-v0}
\label{section_ssf}

The benchmark set that we propose in this subsection demonstrates that in the absence of rewards, QD algorithms can get trapped in some areas of the search space, unless they have the capacity to adapt themselves (\textit{e.g.} their mutations) to the surface's structure. The benchmark, that we have dubbed SSF-v0 for \textbf{S}elf-\textbf{S}imilar \textbf{F}unction \textbf{V}ersion 0, is a set of environments, each of which can be instantiated via a choice of genotype dimensionality. We will call an SSF-v0 problem with genotype dimensionality $N$ an \textit{SSF-V0 problem of order N}. The corresponding behavior mapping will be written $\phi_{ssf}^N$.

Before presenting formal definitions, let us briefly explain the intuition behind the benchmark's design using the SSF-V0 of order 1, shown in figure \ref{fig_self_similar_1} (top). The behavior mapping function $\phi_{ssf}^1$ results from replicating and scaling the same linear and constant parts an infinite number of times. Let us denote by $l_0, l_1, ...$ the series of intervals over which the function is constant (sorted by increasing interval length)\footnote{Since $l_0$ is fixed, the function's graph is not a fractal as it does not cover arbitrarily small scales}. Assume that the population is initialized at the origin. Suppose a QD algorithm has reached an interval $l_i$ at some generation $g$,  \textit{i.e.} that the population $\mathcal{P}^g$ contains a number of individuals $x_1,...,x_l$ that fall in $l_i$. Then, it will only be able to escape that plateau if the mutations that it applies to the $x_i$ are large enough for an offspring at some future generation $\mathcal{P}^{g+t}$ to fall with non-zero probability inside the linear regime that follows $l_i$.

This is illustrated in figure \ref{fig_bench2_a}. The top and bottom plots respectively correspond to NS runs with Gaussian mutations of standard deviation $\sigma=20$ and $\sigma=50$. Both cases show the results after $1000$ generations, and in both situations, the exploration has been blocked at a plateau for more than $500$ generations. This indicates that fixed mutations can only prevent standard QD algorithms from being trapped in areas of constant behavior $l_i$ as long as the length of that interval stays below a certain threshold.

Concretely, we define an environment from that benchmark as follows.

\begin{definition} {SSF-v0 of order $N$.} Given a choice of dimensionality $N$, let the genotype 

\begin{equation}
  \mathcal{G}_{ssf}^N\triangleq \mathbb{R}^N
\end{equation}

and define the sequence $\{R_i\}_i$ as

\begin{equation}
  \begin{cases}
    R_0=0 \\
    R_i=R_{i-1}+2(\floor{\frac{i}{2}})^3 +1  & i=2k+1 \\
    R_i=R_{i-1}+2(\floor{\frac{i}{2}}-1)^3 +1  & i=2k.
  \end{cases}
\end{equation}

\noindent Then, the behavior mapping will be given by

  \begin{equation}
  \begin{cases}
    \phi_{ssf}^N(g)=g & R_{2k}\leq ||g||_2<R_{2k+1}\\
    \phi_{ssf}^N(g)=R_{2k+1} & R_{2k+1}\leq ||g||_2<R_{2(k+1)}.
  \end{cases}
\end{equation}
\end{definition}

\noindent An example of SSF-v0 or order 2 is given in figure \ref{fig_self_similar_1} (bottom).

While it is mathematically clear \footnote{Let us assume that a QD algorithms reaches a given plateau $l_i$, \textit{i.e.} that at least one individual $x$ in its current population is in that interval. In that situation, without any selection mechanism, a purely Gaussian random walk with std $\sigma$ starting at $x$ will result in a distribution $\mathcal{N}(x,\sqrt{n\sigma})$ after $n$ steps. The behavior-based selection pressure applied by the QD algorithm will further skew that distribution towards the linear regime between $l_{i-1}$ and $l_i$. This reduces the likelihood that an individual will reach $l_{i+1}$. Furthermore, the choice of $i$ was arbitrary, and the $l_j$ will grow continuously in length for $j>i$.
} that fixed standard mutations with selection based on behavioral novelty eventually lead to being blocked at one of the plateaus, we ran both NS and MAP-elites experiments with $\sigma=20, 50$ for additional verification. We ran each experiment $20$ times. As expected, none of the instances were able to make any progress beyond the points that we reported previously in figure \ref{fig_bench2_a}. 

A potential solution to the problem that this benchmark highlights is to use adaptive mutations. While such mutations are standard in fitness-driven optimization (\textit{e.g.} CMA-ES\cite{hansen2016cma}), their use in the QD literature remains relatively rare \cite{fontaine2020covariance, paolo2021sparse}, and even then, mutations are only tailored according to reward/fitness, without consideration for behavioral plateaus. Usually, the standard deviation of the mutations is considered as a hyperparameter that is fixed by the practitioner, sometimes via a trial and error process.

\begin{figure}[ht!]
  \includegraphics[width=75mm,trim={2.1cm 0.5cm 1.9cm 0.5},clip]{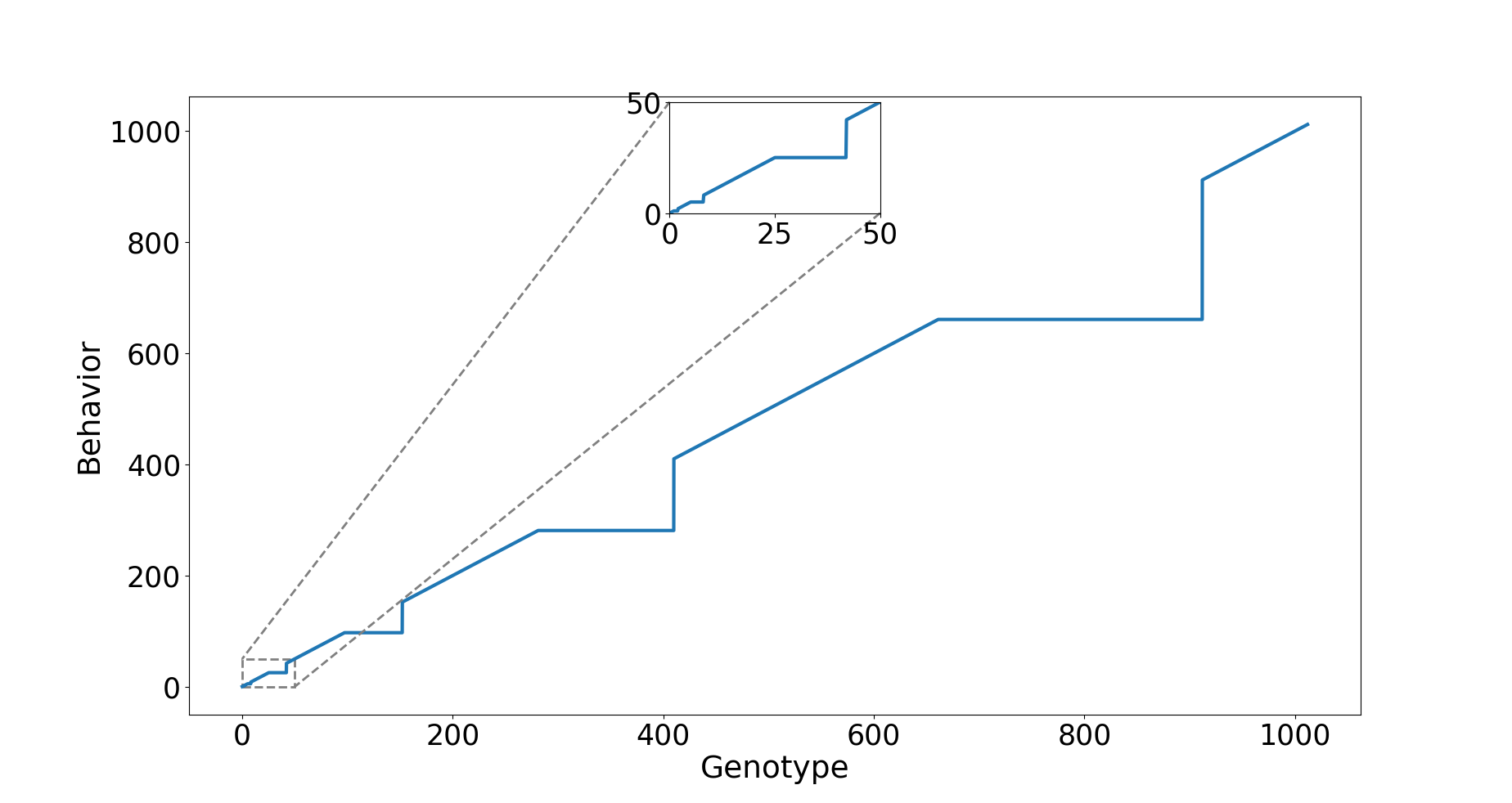}\\
  \includegraphics[width=69mm,trim={2.0cm 0.5cm 1.9cm 1.0},clip]{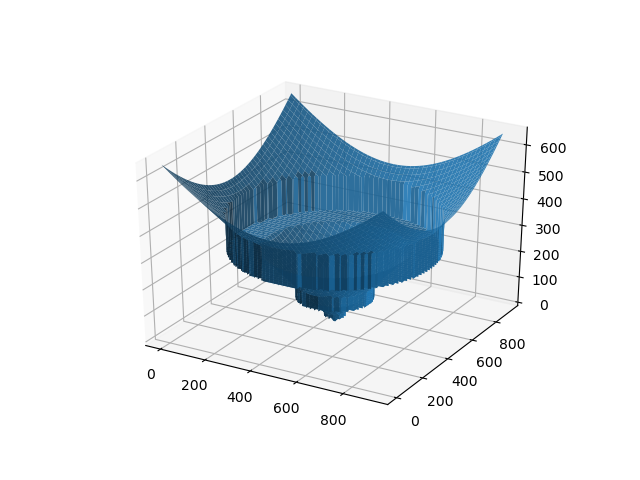}\\
  \caption{\textbf{(Top)} SSF-v0 of order 1. The horizontal axis is the one-dimensional genotype, and the horizontal axis is the behavior space. As highlighted by the zooming window in the figure, the mapping $\phi_{ssf}^1$ results from replicating and rescaling the same linear and constant segments. The function is designed to illustrate that the sequence $l_0, l_1, ...$ of intervals of increasing length over which $\phi_{ssf}^1$ is constant can act as a series of barriers that can not be overcome without adapting the mutation/selection process to the surface's structure. Indeed, taking $l_i$ and $l_j$ with $i<<j$, the \textit{minimum mutation standard deviation} that allows a divergent search algorithm to "escape" from $l_i$ is much smaller than the minimum standard deviation that would allow it to bypass $l_j$. \textbf{(Bottom)} SSF-v0 of order 2.}
  \label{fig_self_similar_1}
\end{figure}

\begin{figure}[ht!]
  \subfloat[Gaussian mutations with std $\sigma=20$.][Gaussian mutations with std $\sigma=20$.]{
      \includegraphics[width=75mm,trim={0cm 0 0cm 0cm},clip]{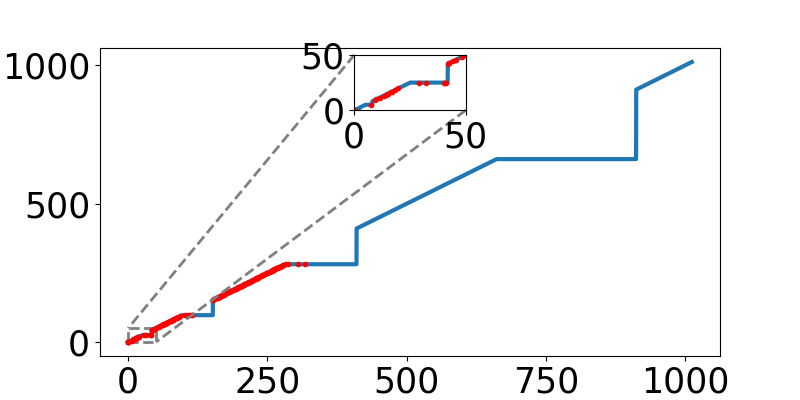}}\\
    \subfloat[Gaussian mutations with std $\sigma=50$.][Gaussian mutations with std $\sigma=50$.]{
      \includegraphics[width=75mm,trim={0cm 0 0cm 0cm},clip]{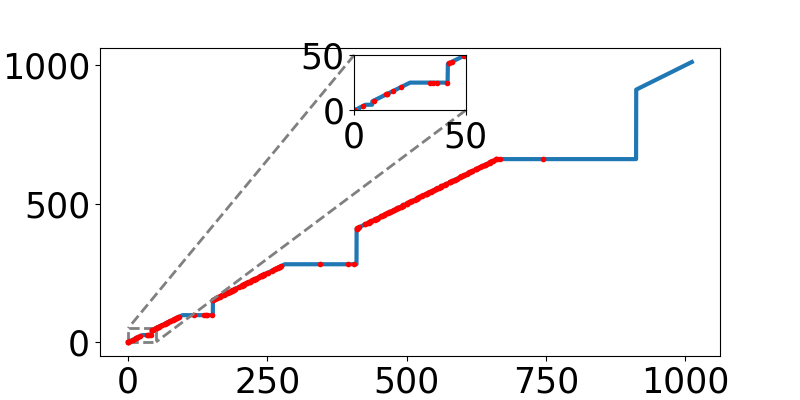}}
  \caption{The results of NS runs with different Gaussian mutation standard deviations. Both cases show the results after $1000$ generations as exploration has been unable to progress past one of the plateaus for about $500$ generations. As the difference between the two plots show, setting $\sigma=50$ only enables the algorithm to escape a single additional plateau compared to the case in which $\sigma=20$.}
  \label{fig_bench2_a}
\end{figure}

\section{Evolvability trap: Deceptive-Evolvability-v0}
\label{section_deceptive_evo}

Current state of the art QD algorithms do not take into account the fact that the mapping $\phi$ from genotypes to behaviors is many to one, and as a result do not discriminate between different solutions that produce similar behaviors. Indeed, two solutions $x_1, x_2 \in \mathcal{G}$ for which $\phi(x_1)=\phi(x_2)$ can exhibit different properties that might make one solution preferable to the other. One possible aspect to consider when choosing between the two is the brittleness of the solution: even in situations where $x_1$ and $x_2$ have the same evolvability, mutating $x_1$ could result in individuals that exhibit novel behaviors but still solve the task at hand, but mutating $x_2$ can result in novel behaviors that are of no interest at all. A more critical problem arises when $x_1$ and $x_2$ differ in terms of evolvability. In that situation, a QD algorithm is likely to focus on the offsprings of the most evolvable one, therefore abandoning areas of the genotype space that could have been more promising in the long term. 

We conjecture that the second situation mentioned above is responsible for poor behavior space coverage in at a number of practical applications \footnote{An example that has motivated this benchmark is robot navigation in complex environments, where two neural network controllers that are both capable of solving the same task $\mathcal{T}_1$, but whose weights lie in different areas of genotype space, exhibit different evolvabilities and thus, do not have the same potential for discovering solutions to tasks $\mathcal{T}_2, ..., \mathcal{T}_n$ that are follow-ups to $\mathcal{T}_1$.}, and propose a simple benchmark problem that models that situation.

Let us define the genotype space as $\mathcal{G}_{de}\triangleq [0, L]\times[0,L]$ with $L \in \mathbb{R}^{+}$, and let us define the behavior function as an unnormalized Gaussian Mixture with two isotropic components:

\begin{equation}
  \phi_{de}(x)=\mathcal{N}(x|\mu_1, \sigma_1I_2) + \beta\mathcal{N}(x|\mu_2, \sigma_2I_2)
\end{equation}

such that $\sigma_1 < \sigma_2$ and $\beta\in \mathbb{R}^{+}$ is sufficiently large to ensure that

\begin{equation}
  \argmax_x \phi_{de}(x)=\mu_2.
\end{equation}

Of particular interest to us are configurations that are similar to the one shown in figure \ref{fig_bench3_a}, where $\sigma_1$ is significantly smaller than $\sigma_2$ and $\beta$ is a relatively large scalar. Let us use the saddle point $x_{saddle}$ (defined by null gradients and a negative Hessian, marked with a red dot in figures \ref{fig_bench3_a} top and bottom left) to initialize the population of the QD algorithm that we want to benchmark, \textit{i.e.} $\mathcal{P}^0 \sim \mathcal{N}(x_{saddle},\sigma)$ where $\sigma$ is the standard deviation of the mutation operator. Gaussian mutations are used in what follows, but extending the discussion to variants of the bounded polynomial operator \cite{deb2002fast} is straight-forward.

Noting $M_1\triangleq \mathcal{N}(\mu_1|\mu_1,\sigma_1I_2)$ and $M_2\triangleq \mathcal{N}(\mu_2|\mu_2,\sigma_2I_2)$, the reachable behavior space can be written

\begin{equation}
  [0,M_2]=[0,M_1]\cup[M_1,M_2].
\end{equation}

While $[0,M_1]$ can be covered by the QD algorithm by "climbing" towards the mode that results from the Gaussian with smaller variance $\sigma_1$, covering the interval $[M_1,M_2]$ requires the exploration to move towards the other mode. Note that in the example of figure \ref{fig_bench3_a}, $[M_1, M_2]$ corresponds to about one third of the reachable behavior space.

\begin{figure}[h!]
  \includegraphics[width=75mm,clip]{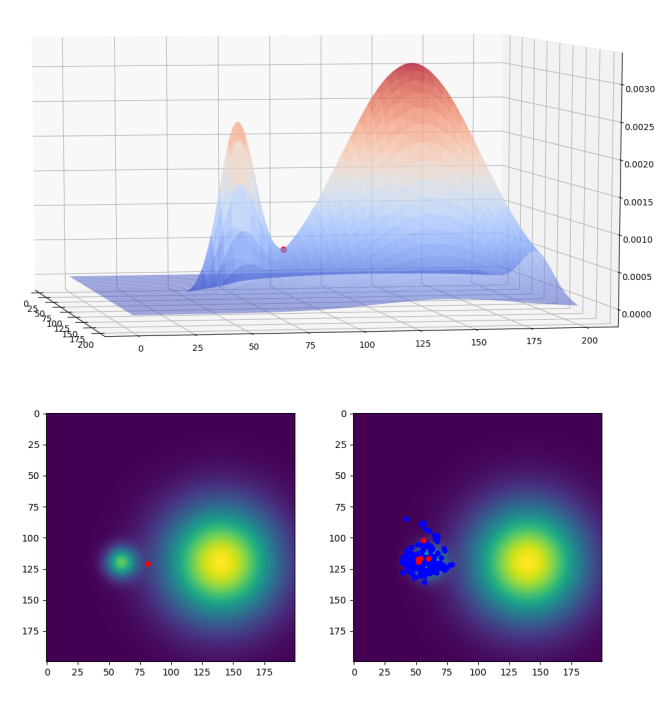}\\
  \caption{(top) 3d view of the proposed benchmark problem, in which $\sigma_1^2=70$ and $\sigma_2^2=1e4$, and $\beta=20$. The red dot indicates the saddle point at which the population is initialized. As explained in the text, individuals falling in the area in which the dominant component of the GMM is the one that has smaller standard deviation $\sigma_1$ will have higher evolvability and thus be more likely to be selected by the QD algorithm. (bottom left) For better visibility, a heatmap view of the problem is provided, with the initialization point also marked as a red dot. (bottom right) Most likely outcome of QD exploration after $100$ generations. Blue dots are archive individuals, and red dots indicate the active population.}
  \label{fig_bench3_a}
\end{figure}

\begin{figure}[h!]
  \centering
      \begin{tikzpicture}
        \node[anchor=south west,inner sep=0] (image) at (0,0)
        {
        \includegraphics[width=75mm,clip]{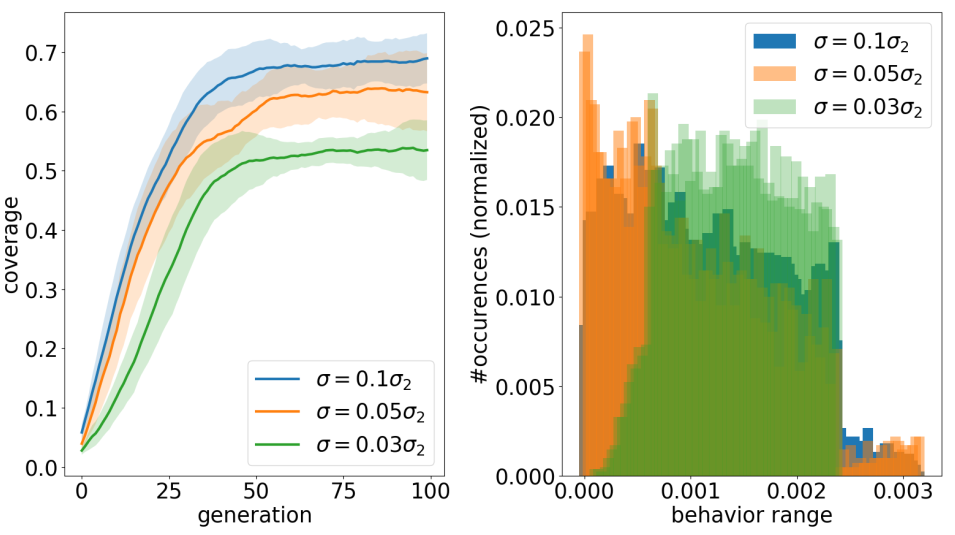}};
        \begin{scope}[x={(image.south east)},y={(image.north west)}]
          \node (A) at (0.875, -0.2) {$M_1$};
          \node (B) at (0.875, 0.096) {};
          \draw[red, thick, ->] (A) edge (B);
          \node (C) at (0.966, -0.2) {$M_2$};
          \node (D) at (0.966, 0.096) {};
          \draw[red, very thick, ->] (C) edge (D);
        \end{scope}
      \end{tikzpicture}
  \caption{Results of NS experiments on the deceptive evolvability problem, for different mutation standard deviations, averaged over $40$ runs for each case. (Left) Evolution of behavior space coverage (mean and std). (Right) Histogram of reached behaviors in the $[0,M_2]$ interval. Due to differences in evolvability, the population that is initialized at the saddle point will most likely converge to the mode that corresponds to the GM component with smaller standard deviation, and therefore, NS runs that discover the $[M_1, M_2]$ behavior interval are rare.}
  \label{fig_bench3_b}
\end{figure}
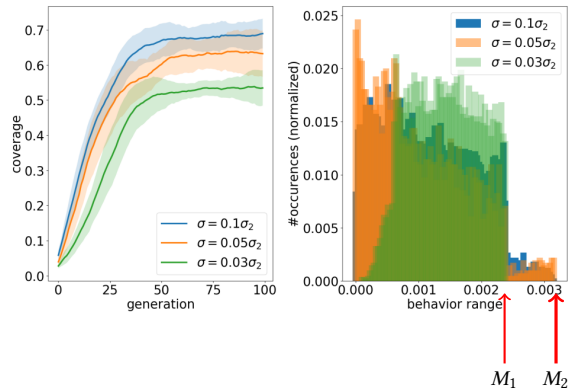

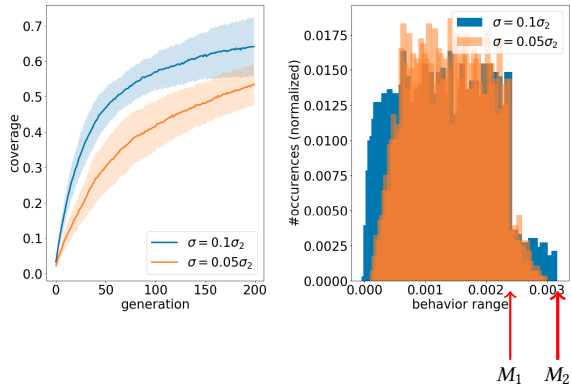
\begin{figure}[h!]
  \centering
      \begin{tikzpicture}
        \node[anchor=south west,inner sep=0] (image) at (0,0)
        {
        \includegraphics[width=75mm,clip]{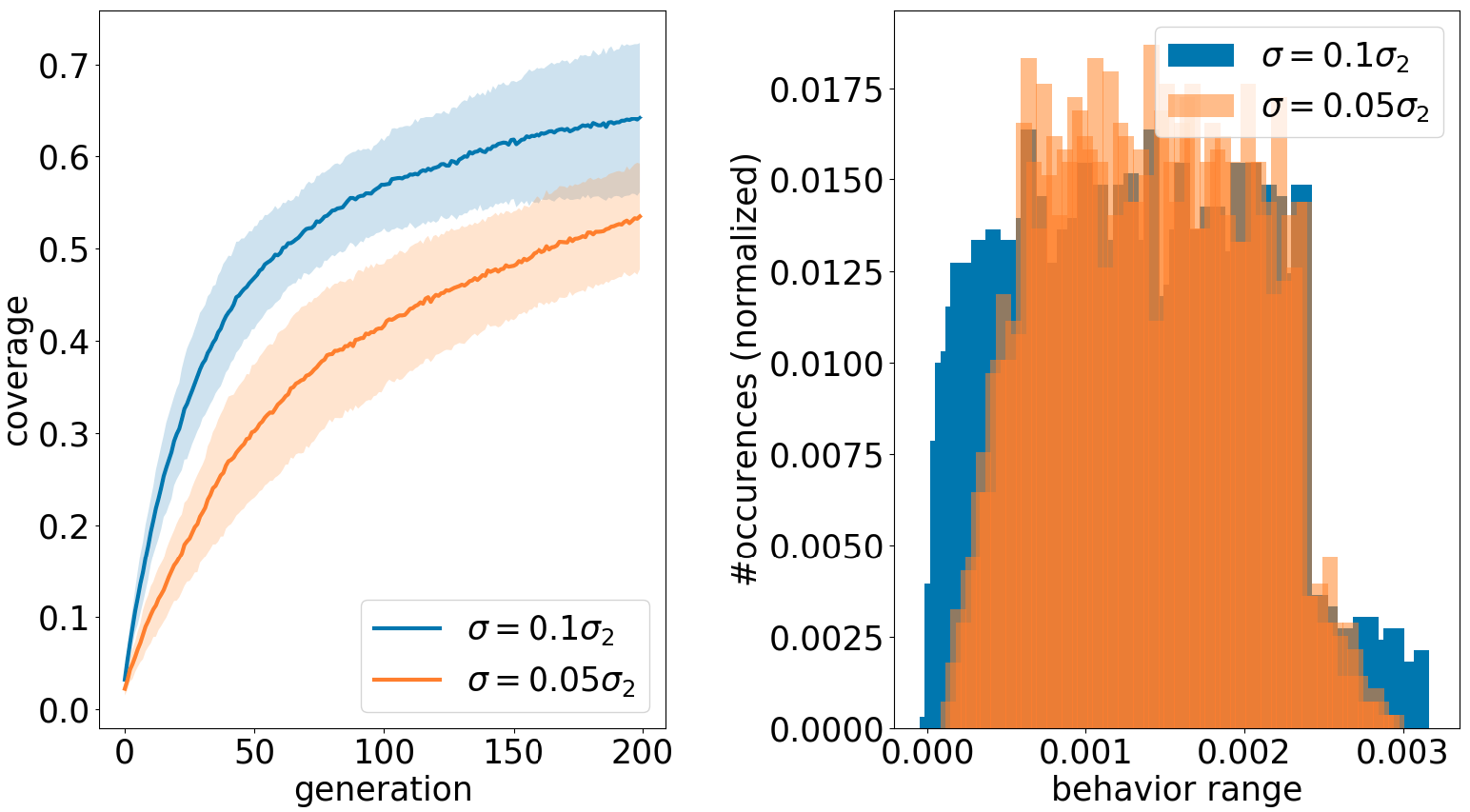}};
        \begin{scope}[x={(image.south east)},y={(image.north west)}]
          \node (A) at (0.888, -0.2) {$M_1$};
          \node (B) at (0.888, 0.096) {};
          \draw[red, thick, ->] (A) edge (B);
          \node (C) at (0.972, -0.2) {$M_2$};
          \node (D) at (0.972, 0.096) {};
          \draw[red, very thick, ->] (C) edge (D);
        \end{scope}
      \end{tikzpicture}
  \caption{Results of MAP-elites experiments for different mutation standard deviations, averaged over $40$ runs for each case. Due to its uniform sampling in the entire archive, MAP-elites converges much more slowly than NS (see figure \ref{fig_bench3_a}). For this reason, we have not included map-elites with $\sigma=3\%\sigma_2$ in that figure. (Left) Evolution of behavior space coverage (mean and std). (Right) Histogram of reached behaviors in the $[0,M_2]$ interval. As in the case of NS, the population that is initialized at the saddle point will most likely converge to the mode that corresponds to the GM component with smaller standard deviation, and therefore, MAP-elites also fails to discover the $[M_1, M_2]$ behavior interval (which corresponds to about one third of the reachable behavior space).}
  \label{fig_bench3_c}
\end{figure}

Since $\phi_{de}$ is smooth and the Gaussians are isotropic, evolvability in that situation reduces to gradient: indeed, the behavior space that is reachable by applying Gaussian mutations $\delta x \sim \mathcal{N}(0, \sigma)$ to an individual $x$ is determined by the gradient of $\phi_{de}$ at $x$. Furthermore, since $\phi_{de}$ is injective in sufficiently small neighborhoods of $x$ there is no difference in uniformity statistics between individuals. In the configuration given in figure \ref{fig_bench3_a}, the gradient is higher around the mode that corresponds to the first component of the mixture, which has a smaller standard deviation $\sigma_1I$. Therefore, offsprings produced by individuals that fall inside that area of genotype space are more likely to be considered novel by the QD algorithm.

As a result of those differences in evolvability, QD algorithms are likely to be lured and get trapped in the area that corresponds to the Gaussian with small standard deviation in the particular problem setting shown in figure \ref{fig_bench3_a}. Figure \ref{fig_bench3_a} (bottom right) shows the most likely outcome that we encountered after $100$ generations of Novelty Search, initialized at the saddle point. In that figure, red points are population individuals, and blue points correspond to the individuals that are stored in Novelty Search's archive (with a maximum size of $200$, and random addition/removal).

We report quantitative results for both Novelty Search and MAP-elites in (respectively) figures \ref{fig_bench3_b} and \ref{fig_bench3_c}. We experimented with mutation standard deviations that were fixed w.r.t the standard deviation of the larger component in the GM. More precisely, we considered $\sigma$ values equivalent to $\{3\%, 5\%, 10\%\}$ of $\sigma_2$ for NS-based experiments. Because of the considerably slower convergence of vanilla MAP-elites, which is due to its random sampling from the entire archive, we only report its performance for $\sigma \in \{5\%\sigma_2, 10\%\sigma_2\}$ and omit the smaller $\sigma=3\%\sigma_2$ value. The results were averaged over $40$ run for each case, in each of which, the population was initialized at the saddle point, as previously detailed. 

As the histogram in figure \ref{fig_bench3_b} (right) as well as the one in \ref{fig_bench3_c} (right) show, QD instances that are able to reach the behavior interval $[M_1,M_2]$ are rare, as the majority of the runs get trapped in the deceptive Novelty optimum. This is also confirmed by the mean and standard deviation of the evolution of archive coverage (figure \ref{fig_bench3_b} (left) and figure \ref{fig_bench3_c} (left)).

\section{Conclusion}
\label{section_conclusion}

In this paper, we presented our initial suggestions and efforts towards developing a standard set of benchmark problems for QD methods. Inspired by \texttt{b-suite}\cite{osband2019behaviour}, we required that those benchmarks should be targeted, simple, fast, challenging and scalable. We identified three challenges in sparse reward settings: \textit{distance metric bias}, \textit{behavioral plateaus} and \textit{evolvability traps}. Our experiments showed that those environments where challenging for both Novelty Search and MAP-elites variants. The list of challenges that we emphasized is far from being exhaustive, and other challenges will naturally need to be identified as the field progresses.

\section*{ACKNOWLEDGMENT}
This work was supported by the European Union's H2020-EU.1.2.2 Research and Innovation Program through FET Project VeriDream under Grant Agreement Number 951992.

\bibliographystyle{ACM-Reference-Format}
\bibliography{sample-bibliography} 

\end{small}
\end{document}